\documentclass[runningheads]{llncs}
\usepackage{graphicx}

\usepackage{tikz}
\usepackage{comment}
\usepackage{amsmath,amssymb} 
\usepackage{color}
\usepackage{multirow}
\usepackage{amssymb}
\usepackage{pifont}
\usepackage{afterpage}
\newcommand{\cmark}{\ding{51}}%
\newcommand{\xmark}{\ding{55}}%

\usepackage[accsupp]{axessibility}  

\usepackage[pagebackref,breaklinks,colorlinks]{hyperref}
\usepackage{orcidlink}

\definecolor{mydarkblue}{rgb}{0,0.08,1}
\definecolor{mydarkgreen}{rgb}{0.02,0.6,0.02}
\definecolor{myred}{rgb}{1.0,0.0,0.0}
\definecolor{mypnk}{rgb}{0.94,0.56,0.6}

\newcommand{\argmax}{\operatornamewithlimits{argmax}}

\begin{document}
\pagestyle{headings}
\mainmatter
\def\ECCVSubNumber{6592}  

\title{Inductive and Transductive Few-Shot Video Classification via Appearance and Temporal Alignments} 

\titlerunning{Few-Shot Video Classification via Appearance and Temporal Alignments}
%
\author{Khoi D. Nguyen\inst{1}\orcidlink{0000-0003-4090-1797}\index{Nguyen, Khoi D.} \and
Quoc-Huy Tran\inst{2}\orcidlink{0000-0003-1396-6544} \and
Khoi Nguyen\inst{1}\orcidlink{0000-0002-9259-420X} \and
Binh-Son Hua\inst{1}\orcidlink{0000-0002-5706-8634} \and
Rang Nguyen\inst{1}\orcidlink{0000-0001-7754-5576}}
\authorrunning{Nguyen et al.}
%
\institute{VinAI Research, Vietnam \and Retrocausal, Inc., USA}
\maketitle

\begin{abstract}
We present a novel method for few-shot video classification, which performs appearance and temporal alignments. In particular, given a pair of query and support videos, we conduct appearance alignment via frame-level feature matching to achieve the appearance similarity score between the videos, while utilizing temporal order-preserving priors for obtaining the temporal similarity score between the videos. Moreover, we introduce a few-shot video classification framework that leverages the above appearance and temporal similarity scores across multiple steps, namely prototype-based training and testing as well as inductive and transductive prototype refinement. To the best of our knowledge, our work is the first to explore transductive few-shot video classification. Extensive experiments on both Kinetics and Something-Something V2 datasets show that both appearance and temporal alignments are crucial for datasets with temporal order sensitivity such as Something-Something V2. Our approach achieves similar or better results than previous methods on both datasets. Our code is available at \url{https://github.com/VinAIResearch/fsvc-ata}.

\keywords{Few-Shot Learning, Video Classification, Appearance Alignment, Temporal Alignment, Inductive Inference, Transductive Inference}
\end{abstract}

\section{Introduction} \label{sec:intro}
Recognizing video contents plays an important role in many real-world applications such as video surveillance~\cite{zhan2008crowd,rodriguez2011data}, anomaly detection~\cite{sultani2018real,haresh2020towards}, video retrieval~\cite{geetha2008survey,snoek2009concept}, and action segmentation~\cite{konin2020retroactivity,khan2022timestamp}. 
In the modern era of deep learning, there exist a large number of studies focusing on learning to classify videos by fully supervising a neural network with a significant amount of labeled data~\cite{tran2015learning,carreira2017quo,wang2018non,tran2018closer}.
While these fully-supervised approaches provide satisfactory results, the high costs of data collection and annotation make it unrealistic to transfer an existing network to new tasks. 
To reduce such high costs, 
few-shot learning~\cite{santoro2016meta,vinyals2016matching,snell2017prototypical,finn2017model,li2017meta,qiao2018few,sung2018learning,rusu2018meta,oreshkin2018tadam} is an emerging trend that aims to adapt an existing network to recognize new classes with limited training data. 
Considerable research efforts have been invested in few-shot learning on images~\cite{wang2021role,Fei_2021_ICCV,le2021poodle,yang2021free,das2021importance,ma2022fewshot,ghaffari2022on}. Extending few-shot learning to videos has been rather limited. 

The main difference between a video and an image is the addition of temporal information in video frames. 
To adapt few-shot learning for videos, recent emphasis has been put on temporal modeling that allows the estimate of the similarity between two videos via frame-to-frame alignment. 
In the few-shot setting, this similarity function is crucial because it helps classify a query video by aligning it with the given support videos.
Early methods~\cite{zhang2020few,bo2020few,fu2020depth,ben2021taen,cao2021few} achieve low performance as they neglect temporal modeling and simply collapse the temporal information in their video representation learning. 
Recent methods~\cite{cao2020few,lu2021few} jointly consider appearance and temporal information by using Dynamic Time Warping~\cite{cuturi2017soft} or Optimal Transport~\cite{cuturi2013sinkhorn} to align the videos.

\begin{figure}[t]
\centering
\includegraphics[width=\linewidth, trim = 2mm 0mm 2mm 0mm,clip]{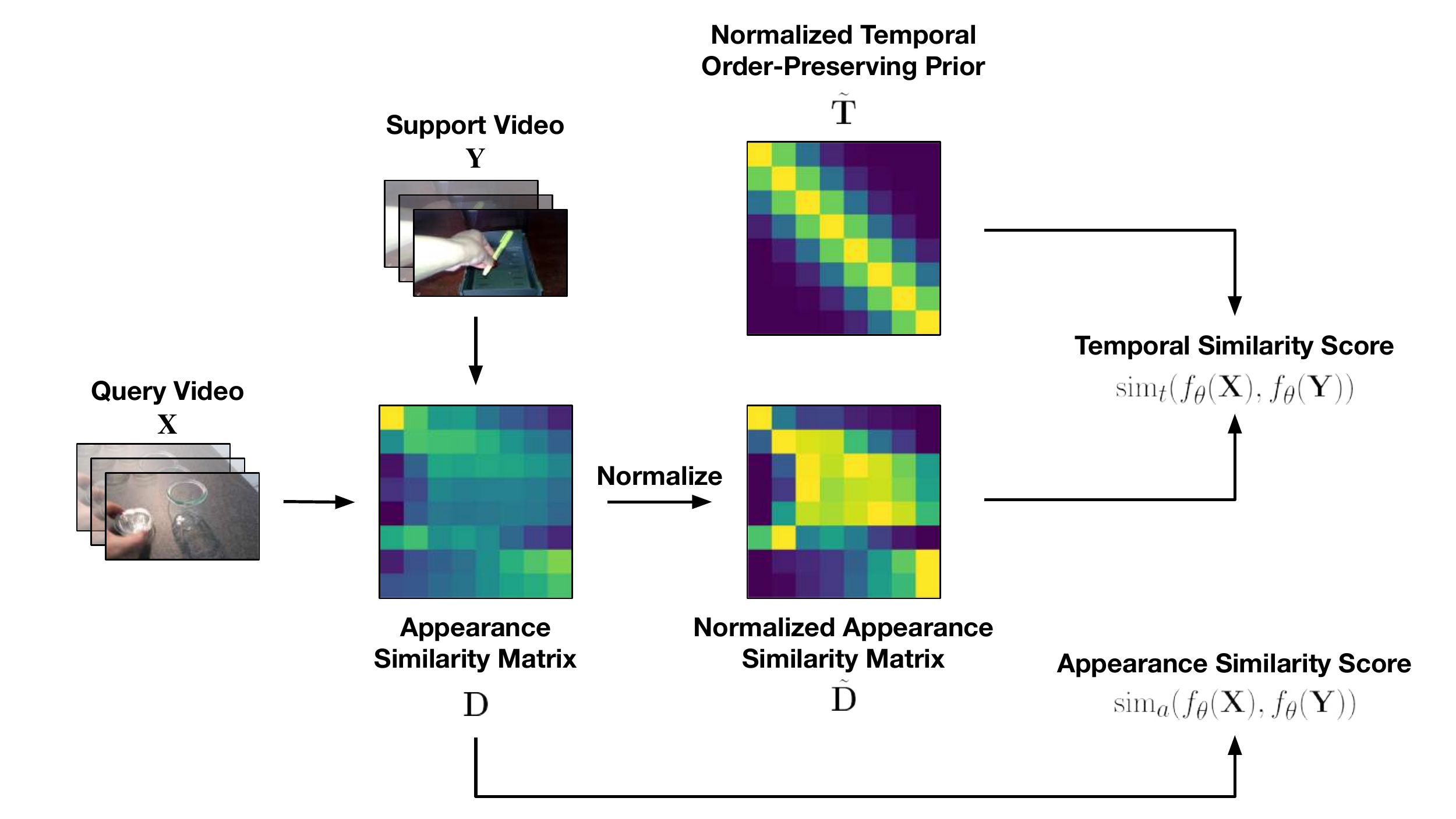}
\caption{\textbf{Our Few-Shot Video Classification Approach.} Given a pair of query and support videos, we first perform appearance alignment (i.e., via frame-level feature matching) to compute the appearance similarity score, and then temporal alignment (i.e., by leveraging temporal order-preserving priors) to calculate the temporal similarity score. 
The final alignment score between the two videos is the weighted sum of appearance and temporal similarity scores.}
\label{fig:teaser}
\end{figure}

We propose to separately consider appearance and temporal alignments, yielding robust similarity functions for use in training and testing in both inductive and transductive few-shot learning.
Following previous works~\cite{cao2020few,zhu2021closer}, we sparsely sample a fixed number of frames from each video and extract their corresponding features using a neural network-based feature extractor. We then compute the pairwise cosine similarity between frame features of the two videos, yielding the appearance similarity matrix. To compute the appearance similarity score between the videos, we first match each frame from one video to the most similar frame in the other, ignoring their temporal order. We then define our appearance similarity score between the two videos as the total similarity of all matched frames between them. Next, motivated by the use of temporal order-preserving priors in different video understanding tasks~\cite{su2017order,cao2020few,haresh2021learning,kumar2021unsupervised} (i.e., initial frames from a source video should be mapped to initial frames in a target video, and similarly the subsequent frames from the source and target video should match accordingly), we encourage the appearance similarity matrix to be as similar as possible to the temporal order-preserving matrix. Our temporal similarity score between the two videos is then computed as the negative Kullback-Leibler divergence between the above matrices. Furthermore, we show how to apply the above appearance and temporal similarity scores in different stages of few-shot video classification, from the prototype-based training and testing procedures to the finetuning of prototypes in inductive and transductive settings. Our method achieves the state-of-the-art performance on both Kinetics~\cite{kay2017kinetics} and Something-Something V2~\cite{goyal2017something} datasets. 
Fig.~\ref{fig:teaser} illustrates our ideas.

In summary, our contributions include:
\begin{itemize}
    \item We introduce a novel approach for few-shot video classification leveraging appearance and temporal alignments. Our main contribution includes an appearance similarity score based on frame-level feature matching and a temporal similarity score utilizing temporal order-preserving priors.
    \item We incorporate the above appearance and temporal similarity scores into various steps of few-shot video classification, from the prototype-based training and testing procedures to the refinement of prototypes in inductive and transductive setups. To our best knowledge, our work is the first to explore transductive few-show video classification.
    \item Extensive evaluations demonstrate that our approach performs on par with or better than previous methods in few-shot video classification on both Kinetics and Something-Something V2 datasets.
\end{itemize}
\section{Related Work}


\noindent \textbf{Few-Shot Learning.} Few-shot learning aims to extract task-level knowledge from seen data while effectively generalizing learned meta-knowledge to unknown tasks. Based on the type of meta-knowledge they capture, few-shot learning techniques can be divided into three groups. Memory-based methods~\cite{santoro2016meta,munkhdalai2019metalearned} attempt to solve few-shot learning by utilizing external memory. Metric-based methods~\cite{vinyals2016matching,snell2017prototypical,qiao2018few,sung2018learning,hou2019cross,wu2021task,kang2021relational,zhang2020deepemd} learn an embedding space such that samples of the same class are mapped to nearby points whereas samples of different classes are mapped to far away points in the embedding space.
Zhang et al.~\cite{zhang2020deepemd} introduce the earth mover’s distance to few-shot learning, which performs spatial alignment between images.
Optimization-based approaches~\cite{finn2017model,li2017meta,rusu2018meta,oreshkin2018tadam,zhang2021shallow} train a model to quickly adapt to a new task with a few gradient descent iterations. One notable work is MAML~\cite{finn2017model}, which learns an initial representation that can effectively be finetuned for a new task with limited labelled data. A new group of works~\cite{le2021poodle,yang2021free,das2021importance} utilize information from meta-training dataset explicitly during meta-testing to boost the performance. While Le et al.~\cite{le2021poodle} and Das et al.~\cite{das2021importance} use base samples as distractors for refining classifiers, Yang et al.~\cite{yang2021free} finds similar features from base data to augment the support set. 

\noindent \textbf{Few-Shot Video Classification.} Zhu and Yang~\cite{zhu2018compound} extend the notion of memory-based meta-learning for video categorization by learning many key-value pairs to represent a video. However, the majority of existing few-shot video classification methods are based on metric-based meta-learning, in which a generalizable video metric is learned from seen tasks and applied to novel unseen tasks. The main difference between a video and an image is the extra temporal dimension of the video frames. Extending few-shot learning to videos thus requires the temporal modeling of this extra dimension. This field of research can be divided in two main groups: aggregation-based and matching-based. The former prioritizes semantic contents, whereas latter is concerned with temporal orderings. Both of these groups start by sampling frames or segments from a video and obtaining their representations using a pretrained encoder. However, aggregation-based methods generate a video-level representation for distance calculation via pooling~\cite{ben2021taen,cao2021few}, adaptive fusion~\cite{bo2020few,fu2020depth}, or attention~\cite{zhang2020few}, whereas matching-based methods explicitly align two sequences and define distance between the videos as their matching cost via different matching techniques such as Dynamic Time Warping (DTW)~\cite{cao2020few} and Optimal Transport (OT)~\cite{lu2021few}. 
While \cite{cao2020few} performs temporal alignment between videos via DTW, which strictly enforces a monotonic frame sequence ordering, \cite{lu2021few} handles permutations using OT with the iterative Sinkhorn-Knopp algorithm, which can be practically slow and poorly scaled~\cite{wertheimer2021few}.
Our proposed method balances between appearance and temporal alignments which allows permutations to some extent. Moreover, it is non-iterative, hence more computationally efficient. Lastly, \cite{cao2020few} and \cite{lu2021few} explore the inductive setting only.

\noindent\textbf{Transductive Inference.} Transductive learning approaches exploit both the labeled training data and the unlabelled testing data to boost the performance. Many previous works~\cite{le2021poodle,boudiaf2020information,hou2019cross,ziko2020laplacian,ren2018meta} explicitly utilize unsupervised information from the query set to augment the supervised information from the support set. With access to more data, transductive inference methods typically outperform their inductive counterparts. However, existing transductive methods are designed mostly for few-shot image classification. In this work, we develop a cluster-based transductive learning approach with a novel assignment function leveraging both temporal and appearance information to address few-shot video classification.
\section{Our Method}

We present, in this section, the details of our approach. We first describe the problem of few-shot video classification in Sec.~\ref{sec:problem}. Next, we introduce our appearance and temporal similarity scores in Sec.~\ref{sec:similarity}, which are then used in our prototype-based training and testing presented in Sec.~\ref{sec:train_test}. Lastly, we discuss how the prototypes are refined in both inductive and transduction settings in Sec.~\ref{sec:refine}.

\subsection{Problem Formulation}
\label{sec:problem}

In few-shot video classification, we are given a base set $\mathcal{D}_{b} = \left \{ \left ( \mathbf{X}_i, y_i \right ) \right \}_{i=1}^{T_{b}}$, where $\mathbf{X}_i$ and $y_i$ denote a video sample and its corresponding class label for $N_{b}$ classes, respectively, while $T_b$ is the number of samples. This base set is used for training a neural network which is subsequently adapted to categorize unseen videos with novel classes. At test time, we are given a set of support videos $\mathcal{D}_{n}^{s} = \left \{ \left ( \mathbf{X}_i, y_i \right ) \right \}_{i=1}^{N \times K}$, where $N$ and $K$ denote the number of novel classes and the number of video samples per novel class, respectively. Note that the novel classes do not overlap with the base classes. The support set provides a limited amount of data to guide the knowledge transfer from the base classes to the novel classes. The goal at test time is to classify the query videos $\mathcal{D}_{n}^{q}$ into one of these novel classes. Such configuration is called an $N$-way $K$-shot video classification task. In this paper, we explore two configurations: 5-way 1-shot and 5-way 5-shot video classification.

There are two settings for few-shot learning: inductive and transductive learning. In the former, each of the query videos are classified independently, whereas, in the latter, the query videos are classified collectively, allowing unlabeled visual cues to be shared and leveraged among the query videos, which potentially improves the overall classification results. In this work, we consider both inductive and transductive settings. To our best knowledge, our work is the first to explore transductive learning for few-shot video classification.

Following prior works~\cite{cao2020few,zhu2021closer}, we represent a video by a fixed number of $M$ frames randomly sampled from equally separated $M$ video segments, or $\mathbf{X} = [\mathbf{x}^1, \dots, \mathbf{x}^M]$ with $\mathbf{x}^i \in \mathbb{R}^{3\times H \times W}$ and $i \in \{1, \dots, M\}$, while $H$ and $W$ are the width and height of the video frame, respectively. Next, a neural network $f$ with parameters $\theta$ is used to extract a feature vector $f_\theta(\mathbf{x}^i) \in \mathbb{R}^{C}$ ($C$ is the number of channels) for each sampled frame $\mathbf{x}^i$. Lastly, the features of a video $\mathbf{X}$ is represented by $f_\theta(\mathbf{X}) =[f_\theta(\mathbf{x}^1), \dots, f_\theta(\mathbf{x}^M)] \in \mathbb{R}^{C \times M}$.

\subsection{Appearance and Temporal Similarity Scores}
\label{sec:similarity}

\begin{figure}[t]
\centering
\includegraphics[width=\linewidth, trim = 5mm 25mm 38mm 0mm,clip]{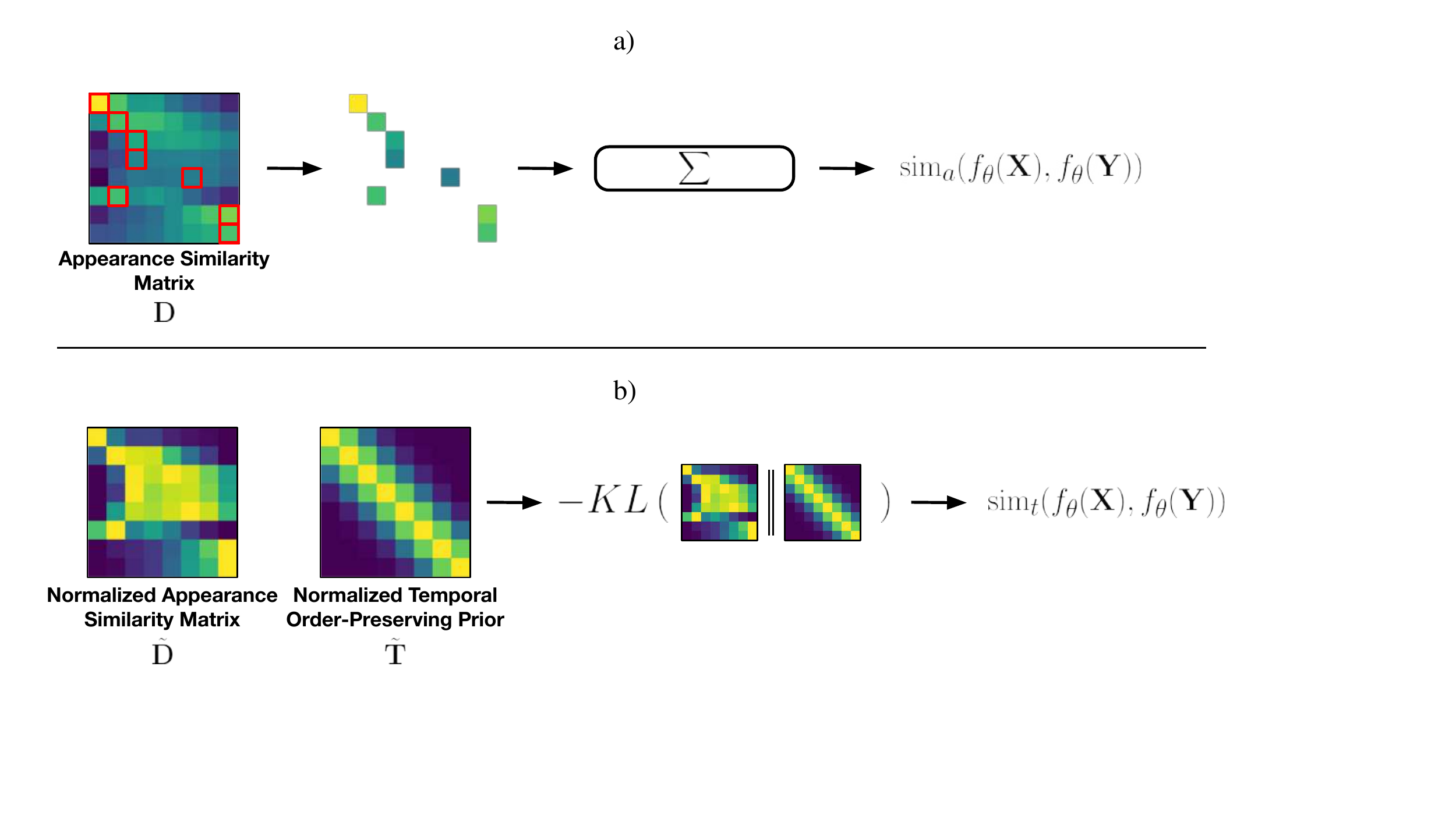}
\caption{\textbf{Appearance and Temporal Similarity Scores.} a) The proposed appearance similarity score is computed as the sum of the frame-level maximum appearance similarity scores between frames in $\mathbf{X}$ and frames in $\mathbf{Y}$. b) The proposed temporal similarity score is based on the Kullback-Leibler divergence between the row-wise normalized appearance similarity matrix $\tilde{\mathbf{D}}$ and the row-wise normalized temporal order-preserving prior $\tilde{\mathbf{T}}$.}
\label{fig:similarity}
\end{figure}

Existing distance/similarity functions are either computationally inefficient~\cite{lu2021few}, imposing too strong constraint~\cite{cao2020few}, or oversimplified that they neglect temporal information~\cite{zhu2021closer}. To capture both appearance and temporal information with low computational costs, we propose to explore appearance and temporal cues via two simple yet novel similarity functions. The final prediction is then a linear combination of the predictions from the two functions. We detailed our similarity functions, below.

\noindent \textbf{Appearance Similarity Score.} Given a pair of videos $(\mathbf{X}, \mathbf{Y})$, we first compute the appearance similarity matrix $\mathbf{D} \in \mathbb{R}^{M \times M}$ between them. The element $\mathbf{D}(i,j)$ of $\mathbf{D}$ is the pairwise cosine similarity between frame $\mathbf{x}^i$ in  $\mathbf{X}$ with frame $\mathbf{y}^j$ in $\mathbf{Y}$ as follows:
\begin{equation}
    \mathbf{D}(i, j) = \frac{f_\theta(\mathbf{x}^i)^T f_\theta(\mathbf{y}^j)}{||f_\theta(\mathbf{x}^i)||~ ||f_\theta(\mathbf{y}^j)||}.
\end{equation}

For every frame $\mathbf{x}^i$ in $\mathbf{X}$, we can align it with the frame $\mathbf{y}^k$ in $\mathbf{Y}$ which has the highest appearance similarity score with $\mathbf{x}^i$ (i.e., $k = \argmax_j \mathbf{D}(i,j)$), ignoring their relative ordering. We define the appearance similarity score between $\mathbf{X}$ and $\mathbf{Y}$ as the sum of the optimal appearance similarity scores for all frames in $\mathbf{X}$ as:
\begin{equation}
\begin{split}
    \text{sim}_{a} (f_\theta(\mathbf{X}), f_\theta(\mathbf{Y})) &= \sum_{i=1}^M \max_j \mathbf{D}(i,j) \approx \sum_{i=1}^M \lambda \log \sum_{j=1}^M \exp^{\frac{\mathbf{D}(i, j)}{\lambda}},
\end{split}
\label{eq:log-sum-exp}
\end{equation}
where we use \emph{log-sum-exp} (with the smoothing temperature $\lambda = 0.1$) to continuously approximate the \emph{max} operator.
Intuitively, the $\text{sim}_{a} (f_\theta(\mathbf{X}),f_\theta(\mathbf{Y}))$ shows how similar in appearance frames in $\mathbf{X}$ to frames in $\mathbf{Y}$.
Note that $\text{sim}_{a}$ is not a symmetric function. Fig.~\ref{fig:similarity}(a) shows the steps to compute our appearance similarity score.

\noindent \textbf{Temporal Similarity Score.} Temporal order-preserving priors have been employed in various video understanding tasks such as sequence matching~\cite{su2017order}, few-shot video classification~\cite{cao2020few}, video alignment~\cite{haresh2021learning}, and activity segmentation~\cite{kumar2021unsupervised}. In particular, given two videos $\mathbf{X}$ and $\mathbf{Y}$ of the same class, it encourages initial frames in $\mathbf{X}$ to be aligned with initial frames in $\mathbf{Y}$, while subsequent frames in $\mathbf{X}$ are encouraged to be aligned with subsequent frames in $\mathbf{Y}$. Mathematically, it can be modeled by a 2D distribution $\mathbf{T} \in \mathbb{R}^{M \times M}$, whose marginal distribution along any line perpendicular to the diagonal is a Gaussian distribution centered at the intersection on the diagonal, as:
\begin{align}
    \mathbf{T}(i,j) = \frac{1}{\sigma \sqrt{2\pi}} \exp^{-\frac{l^2(i,j)}{2\sigma^2}},~~~l(i,j) = \frac{|i - j|}{\sqrt{2}},
\end{align}
where $l(i, j)$ is the distance from the entry $(i, j)$ to the diagonal line and the standard deviation parameter $\sigma = 1$. The values of $\mathbf{T}$ peak on the diagonal and gradually decrease along the direction perpendicular to the diagonal.

In this work, we adopt the above temporal order-preserving prior for few-shot video classification. Let $\tilde{\mathbf{D}}$ and $\tilde{\mathbf{T}}$ denote the row-wise normalized version of $\mathbf{D}$ and $\mathbf{T}$ respectively, as:
\begin{align}
\tilde{\mathbf{D}}(i,j) = \frac{\exp^{\mathbf{D}(i,j)}}{\sum_{k=1}^M \exp^{\mathbf{D}(i,k)}},~~~\tilde{\mathbf{T}}(i,j) = \frac{\mathbf{T}(i,j)}{\sum_{k=1}^M \mathbf{T}(i,k)}. 
\end{align}
We define the temporal similarity score between $\mathbf{X}$ and $\mathbf{Y}$ as the negative Kullback-Leibler (KL) divergence between $\tilde{\mathbf{D}}$ and $\tilde{\mathbf{T}}$:
\begin{align}
    \text{sim}_{t} (f_\theta(\mathbf{X}), f_\theta(\mathbf{Y})) = -KL(\tilde{\mathbf{D}}||\tilde{\mathbf{T}}) =  -\frac{1}{M} \sum_{i=1}^M \sum_{j=1}^M \tilde{\mathbf{D}}(i,j) \log \frac{\tilde{\mathbf{D}}(i,j)}{\tilde{\mathbf{T}}(i,j)}.
\end{align}
Intuitively, $\text{sim}_{t} (f_\theta(\mathbf{X}), f_\theta(\mathbf{Y}))$ encourages the temporal alignment between $\mathbf{X}$ and $\mathbf{Y}$ to be as similar as possible to the temporal order-preserving alignment $\mathbf{T}$. Fig.~\ref{fig:similarity}(b) summarizes the steps to compute our temporal similarity score.

\subsection{Training and Testing}
\label{sec:train_test}

\noindent \textbf{Training.} We employ the global training with prototype-based classifiers~\cite{snell2017prototypical} in our approach. A set of prototypes $\mathbf{W} = [\mathbf{W}_1, \dots, \mathbf{W}_{N_b}]$ are initialized randomly, where each prototype $\mathbf{W}_p = [\mathbf{w}^1_p, \dots, \mathbf{w}^M_p] \in \mathbb{R}^{C \times M}$ represents a class in the base set. Note that in contrast to few-shot image classification~\cite{snell2017prototypical}, where each prototype is a single feature vector, in our approach for few-shot video classification, each prototype is a sequence of $M$ feature vectors. To learn $\mathbf{W}$, we adopt the below supervised loss:
\begin{align}
    \mathcal{L}_{Sup} &= - \mathbb{E}_{(\mathbf{X}, y) \sim \mathcal{D}_b} \log \frac{\exp^{\text{sim}_{a} (f_\theta(\mathbf{X}), \mathbf{W}_y))}}{\sum_{p=1}^{N_b} \exp^{\text{sim}_{a} (f_\theta(\mathbf{X}), \mathbf{W}_p))}} 
     - \alpha \mathbb{E}_{(\mathbf{X}, y) \sim \mathcal{D}_b}
    \text{sim}_{t} (f_\theta(\mathbf{X}), \mathbf{W}_y).
\label{eq:sup_loss}
\end{align}
Here, $\alpha$ is the balancing weight between the two terms, and the appearance and temporal similarity scores are computed between the features of a video and the prototype of a class. We empirically observe that the temporal order-preserving prior is effective for Something-Something V2 but not much for Kinetics. This is likely because the actions in Something-Something V2 are order-sensitive, whereas those in Kinetics are not. Therefore, we use both terms in Eq.~\ref{eq:sup_loss} for training on Something-Something V2 (i.e., $\alpha = 0.05$) but only the first term for training on Kinetics (i.e., $\alpha = 0$).

Solely training with the above supervised loss can lead to a trivial solution that the model only learns discriminative features for each class (i.e., the $M$ feature vectors within each prototype are similar). To avoid such cases, we add another loss which minimizes the entropy of $\tilde{\mathbf{D}}$:
\begin{equation}
\begin{split}
    \mathcal{L}_{Info} &= - \frac{1}{M} \sum_{i=1}^M \sum_{j=1}^M \tilde{\mathbf{D}}(i,j) \log \tilde{\mathbf{D}}(i,j).
\end{split}
\end{equation}

The final training loss is a combination of the above losses and is written as:
\begin{equation}
\begin{split}
\label{loss}
    \mathcal{L} = \mathcal{L}_{Sup} + \nu \mathcal{L}_{Info}.
\end{split}
\end{equation}
Here, $\nu = 0.1$ is the balancing weight.

\noindent \textbf{Testing.} At test time, we discard the global prototype-based classifiers, and keep the feature extractor. Given an $N$-way $K$-shot episode, we first extract the features of all support and query samples. We then initialize $N$ prototypes by the average features of support samples from the corresponding classes, i.e., $\mathbf{W}_c = \frac{1}{K} \sum_{\mathbf{X} \in \mathcal{S}_c} f_\theta(\mathbf{X})$, where $\mathcal{S}_c$ is the support set of the $c$-th class. Given the prototypes, we define the predictive distribution over classes of each video sample $p(c|\mathbf{X}, \mathbf{W}), c \in \{1, \dots, N\}$, as:

\begin{equation}
\begin{split}
    p(c|\mathbf{X}, \mathbf{W}) = (1 - \beta) \frac{\exp^{\text{sim}_{a} (f_\theta(\mathbf{X}), \mathbf{W}_c)}}{\sum_{j=1}^{N} \exp^{\text{sim}_{a} (f_\theta(\mathbf{X}), \mathbf{W}_j)}} + \beta \frac{\exp^{\text{sim}_{t} (f_\theta(\mathbf{X}), \mathbf{W}_c)}}{\sum_{j=1}^{N} \exp^{\text{sim}_{t} (f_\theta(\mathbf{X}), \mathbf{W}_j)}},
\end{split}
\label{eq:pred_dist}
\end{equation}
where $\beta$ is the balancing weight between the two terms. The above predictive distribution is a combination of the \emph{softmax} predictions based on appearance and temporal similarity scores. As we discussed previously, since the actions on Something-Something V2 are order-sensitive but those on Kinetics are not, the temporal order-preserving is effective for Something-Something V2 but not for Kinetics. Therefore, we set $\beta = 0.5$ for Something-Something V2 and $\beta = 0$ for Kinetics. As we empirically show in Sec.~\ref{sec:ablation}, these settings yield good results.

\subsection{Prototype Refinement}
\label{sec:refine}

The prototypes can be further refined with support samples (in the inductive setting) or with both support and query samples (in the transductive setting) before being used for classification.

\noindent \textbf{Inductive Setting.} For the inductive inference, we finetune the prototypes on the support set with the following cross-entropy loss:
\begin{align}
    \mathcal{L}_{inductive} &= - \mathbb{E}_{(\mathbf{X}, y) \sim \mathcal{D}^s} \log \frac{\exp^{\text{sim}_{a} (f_\theta(\mathbf{X}), \mathbf{W}_y)}}{\sum_{i=1}^{N} \exp^{\text{sim}_{a} (f_\theta(\mathbf{X}), \mathbf{W}_i)}}.
\label{eq:induc_refine}
\end{align}

\noindent \textbf{Transductive Setting.} We introduce a transductive inference step that utilizes unsupervised information from the query set to finetune the prototypes. The transductive inference typically has a form of Soft $K$-means~\cite{ren2018meta} with a novel assignment function. For each update iteration, we refine the prototypes with the weighted sums of the support and query samples:

\begin{equation}
\begin{split}
    \mathbf{W}_c = \frac{\sum_{\mathbf{X} \sim \mathcal{S}} f_{\theta}(\mathbf{X}) z(\mathbf{X},c) + \sum_{\mathbf{X} \sim \mathcal{Q}} f_{\theta}(\mathbf{X}) z(\mathbf{X},c)} {\sum_{\mathbf{X} \sim \mathcal{S}} z(\mathbf{X},c) + \sum_{\mathbf{X} \sim \mathcal{Q}} z(\mathbf{X},c)},
\end{split}
\end{equation}
where $\mathcal{S}$ and $\mathcal{Q}$ are the support and query sets, respectively. The assignment function $z(\mathbf{X}, c)$ simply returns the $c$-th element of the one-hot label vector of $\mathbf{X}$ if the sample is from the support set. For query samples, we directly use the predictive distribution in Eq.~\ref{eq:pred_dist}, i.e., $z(\mathbf{X}, c) = p(c|\mathbf{X}, \mathbf{W})$ for $\mathbf{X} \in \mathcal{Q}$. In our experiments, the prototypes are updated for 10 iterations.
\section{Experiments}
There are two parts of experiments in this section. In the first part (Sec.~\ref{sec:ablation}), we conduct ablation studies to show the effectiveness of appearance and temporal similarity scores. In the second part (Sec.~\ref{sec:comparison}), we compare our proposed method to state-of-the art methods on two widely used datasets: Kinetics~\cite{kay2017kinetics}, and Something-Something V2~\cite{goyal2017something}. The evaluations are conducted on both inductive and transductive settings. 

\noindent \textbf{Datasets.} We perform experiments on two standard few-shot video classification datasets: the few-shot versions of Kinetics~\cite{kay2017kinetics} and Something-Something V2~\cite{goyal2017something}. Kinetics~\cite{kay2017kinetics} contains 10,000 videos, while Something-Something V2~\cite{goyal2017something} has 71,796 videos. These two datasets are split into 64 training classes, 16 validation classes, and 20 testing classes, following the splits from~\cite{zhu2018compound} and~\cite{cao2020few}.

\setlength{\tabcolsep}{4pt}
\begin{table}
\begin{center}
\caption{Comparison to different aggregation methods in the \textbf{inductive} setting on the Kinetics and Something-Something V2 datasets.}
\label{tab:matching}
\begin{tabular}{c|c|c|c|c}
\hline\noalign{\smallskip}
 & \multicolumn{2}{c|}{\textbf{Kinetics}} & \multicolumn{2}{c}{\textbf{Something V2}} \\
\textbf{Method} & \textbf{1-shot} & \textbf{5-shot} & \textbf{1-shot} & \textbf{5-shot}\\
\noalign{\smallskip}
\hline
\noalign{\smallskip}
OT & ${61.71} \pm 0.41$ & ${77.16} \pm 0.37$ & $35.60 \pm 0.41$ & $47.82 \pm 0.44$ \\
Max & ${73.46} \pm 0.38$ & $\bf{87.67} \pm 0.29$ & $40.97 \pm 0.41$ & $58.77 \pm 0.43$ \\
\bf{Ours} & $\bf{74.26} \pm 0.38$ & ${87.40} \pm 0.30$ & $\bf{43.82} \pm 0.42$ & $\bf{61.07} \pm 0.42$ \\
\noalign{\smallskip}
\hline
\end{tabular}
\end{center}
\end{table}
\setlength{\tabcolsep}{1.4pt}

\noindent \textbf{Implementation Details.} For fair comparisons, we follow the preprocessing steps from prior works~\cite{cao2020few,zhu2021closer,zhang2021learning}. In particular, we first resize the frames of a particular video to $256 \times 256$ and perform random cropping (for training) or central cropping (for testing) a $224 \times 224$ region from the frames. The number of sampled segments or frames for each video is $M=8$. We use the same network architecture as previous works, which is a ResNet-50 pretrained on ImageNet~\cite{ILSVRC15}. Stochastic gradient descent (SGD) with a momentum of 0.9 is adopted as our optimizer. The model is trained for 25 epochs with an initial learning rate of 0.001 and a weight decay of 0.1 at epoch 20. In the inference stage, we report mean accuracy with $95\%$ confidence interval on 10,000 random episodes.

\subsection{Ablation Study}
\label{sec:ablation}

\noindent \textbf{The Effectiveness of Appearance and Temporal Similarity Scores}.
In this experiment, we compare our appearance and temporal similarity scores with several methods for aggregating the appearance similarity matrix $\mathbf{D}$ in Eq.~\ref{eq:log-sum-exp}, namely the optimal transport (OT) and the max operator. For OT, we use the negative appearance similarity matrix $-\mathbf{D}$ as the cost matrix and employ the equal partition constraint. It optimally matches frames of two videos without considering temporal information. For the max operator, we replace the log-sum-exp operator in Eq.~\ref{eq:log-sum-exp} with the max operator. Tab.~\ref{tab:matching} shows their results on Kinetics and Something-Something V2 datasets. Our method outperforms the other two similarity functions by large margins on both datasets with the exception that the max variant performs slightly better than our method in the Kinetics 5-way 5-shot setting ($87.67\%$ as compared to $87.40\%$ of our method).

\setlength{\tabcolsep}{4pt}
\begin{table}
\begin{center}
\caption{Comparison of different training losses in the \textbf{inductive} and \textbf{transductive} settings on the Kinetics and Something-Something V2 datasets.}
\label{tab:losses}
\begin{tabular}{ccc|c|c|c|c}
\hline\noalign{\smallskip}
 & & & \multicolumn{2}{c|}{\textbf{Kinetics}} & \multicolumn{2}{c}{\textbf{Something V2}} \\
$\mathcal{L}_{Sup}$ & $\mathcal{L}_{Info}$ & Transd. & \textbf{1-shot} & \textbf{5-shot} & \textbf{1-shot} & \textbf{5-shot}\\
\noalign{\smallskip}
\hline
\noalign{\smallskip}
\cmark & & & ${74.25} \pm 0.38$ & ${87.15} \pm 0.30$ & ${41.84} \pm 0.42$ & ${57.26} \pm 0.43$ \\
\cmark & \cmark & & ${74.26} \pm 0.38$ & ${87.40} \pm 0.30$ & ${43.82} \pm 0.42$ & ${61.07} \pm 0.42$ \\
\cmark & & \cmark & ${82.71} \pm 0.44$ & ${92.91} \pm 0.31$ & ${47.02} \pm 0.54$ & ${67.56} \pm 0.56$ \\
\cmark & \cmark & \cmark & $\bf{83.08} \pm 0.44$ & $\bf{93.33} \pm 0.30$ & $\bf{48.67} \pm 0.54$ & $\bf{69.42} \pm 0.55$ \\
\noalign{\smallskip}
\hline
\end{tabular}
\end{center}
\end{table}
\setlength{\tabcolsep}{1.4pt}

\noindent \textbf{The Effectiveness of Training Losses}.
We perform analysis of our training losses in Sec.~\ref{sec:train_test} on both Kinetics and Something-Something V2 datasets.
Results are shown in Tab.~\ref{tab:losses}.
As mentioned earlier, adding $\mathcal{L}_{Info}$ in the training phase helps avoiding trivial solutions and hence consistently improves the performance on both Kinetics and Something-Something V2 datasets. 
Using the complete training loss with the transductive setting yields the best performance on both Kinetics and Something-Something V2 datasets. 

\noindent \textbf{The Relative Importance of Appearance and Temporal Cues}. In this experiment, we investigate the effectiveness of the hyperparameter $\beta$ in the inductive and transductive settings. More specifically, we train our model with the loss in Eq.~\ref{eq:sup_loss} and evaluate the result on the validation set with different values of $\beta$. The results of the inductive and transductive settings are shown in Tab.~\ref{table:abla:inductive}. 

\setlength{\tabcolsep}{4pt}
\begin{table}[h]
\begin{center}
\caption{Ablation study on the relative importance of the appearance and temporal terms in computing the predictive distribution and the assignment function of the \textbf{inductive} and \textbf{transductive} inference on the Kinetics and Something-Something V2 datasets.}
\label{table:abla:inductive}
\begin{tabular}{c|c|c|c|c|c}
\hline\noalign{\smallskip}
 & & \multicolumn{2}{c|}{\textbf{Kinetics}} & \multicolumn{2}{c}{\textbf{Something V2}} \\
$\beta$ & \textbf{Transd.} & \textbf{1-shot} & \textbf{5-shot} & \textbf{1-shot} & \textbf{5-shot}\\
\noalign{\smallskip}
\hline
\noalign{\smallskip}
$1.00$ & \multirow{11}{*}{\xmark} & ${21.09} \pm 0.34$ & ${27.00} \pm 0.39$ & ${25.28} \pm 0.37$ & ${33.57} \pm 0.41$ \\
$0.90$ & & ${24.02} \pm 0.36$ & ${41.75} \pm 0.44$ & ${28.09} \pm 0.38$ & ${43.21} \pm 0.43$ \\
$0.80$ & & ${28.94} \pm 0.38$ & ${62.66} \pm 0.43$ & ${32.24} \pm 0.40$ & ${54.45} \pm 0.43$ \\
$0.70$ & & ${39.75} \pm 0.42$ & ${79.99} \pm 0.35$ & ${39.13} \pm 0.42$ & ${62.73} \pm 0.42$ \\
$0.60$ & & ${58.87} \pm 0.43$ & ${86.37} \pm 0.30$ & ${45.40} \pm 0.43$ & ${65.99} \pm 0.41$ \\
$0.50$ & & ${74.05} \pm 0.37$ & ${87.33} \pm 0.29$ & ${48.05} \pm 0.43$ & $\bf{66.91} \pm 0.40$ \\
$0.40$ & & ${75.16} \pm 0.37$ & $\bf{87.38} \pm 0.29$ & $\bf{48.55} \pm 0.42$ & ${66.83} \pm 0.40$ \\
$0.30$ & & ${75.34} \pm 0.36$ & ${87.38} \pm 0.29$ & ${48.54} \pm 0.42$ & ${66.64} \pm 0.41$ \\
$0.20$ & & ${75.37} \pm 0.36$ & ${87.37} \pm 0.29$ & ${48.36} \pm 0.43$ & ${66.32} \pm 0.41$ \\
$0.10$ & & $\bf{75.38} \pm 0.36$ & ${87.36} \pm 0.29$ & ${48.21} \pm 0.43$ & ${66.06} \pm 0.41$ \\
$0.00$ & & ${75.37} \pm 0.36$ & ${87.35} \pm 0.29$ & ${48.13} \pm 0.43$ & ${65.77} \pm 0.41$ \\
\noalign{\smallskip}
\hline\noalign{\smallskip}
\noalign{\smallskip}
$1.00$ & \multirow{9}{*}{\cmark} & ${54.83} \pm 0.49$ & ${66.61} \pm 0.50$ & ${39.53} \pm 0.48$ & ${51.46} \pm 0.51$ \\
$0.90$ & & ${63.88} \pm 0.49$ & ${80.48} \pm 0.43$ & ${43.00} \pm 0.50$ & ${58.45} \pm 0.53$ \\
$0.80$ & & ${72.38} \pm 0.48$ & ${88.71} \pm 0.36$ & ${46.56} \pm 0.53$ & ${65.33} \pm 0.54$ \\
$0.70$ & & ${78.09} \pm 0.45$ & ${91.49} \pm 0.32$ & ${50.03} \pm 0.54$ & ${70.46} \pm 0.53$ \\
$0.60$ & & ${80.73} \pm 0.44$ & ${92.66} \pm 0.31$ & ${52.36} \pm 0.55$ & ${73.42} \pm 0.52$ \\
$0.50$ & & ${82.34} \pm 0.43$ & ${93.24} \pm 0.30$ & ${53.65} \pm 0.55$ & ${74.77} \pm 0.52$ \\
$0.40$ & & ${83.29} \pm 0.43$ & ${93.51} \pm 0.29$ & ${54.46} \pm 0.55$ & ${75.42} \pm 0.52$ \\
$0.30$ & & ${83.81} \pm 0.43$ & ${93.59} \pm 0.29$ & $\bf{54.67} \pm 0.55$ & $\bf{75.47} \pm 0.52$ \\
$0.20$ & & ${84.06} \pm 0.43$ & $\bf{93.60} \pm 0.29$ & ${54.75} \pm 0.55$ & ${75.25} \pm 0.52$ \\
$0.10$ & & ${84.33} \pm 0.42$ & ${93.55} \pm 0.29$ & ${54.66} \pm 0.55$ & ${74.77} \pm 0.52$ \\
$0.00$ & & $\bf{84.36} \pm 0.41$ & ${93.42} \pm 0.29$ & ${54.4} \pm 0.54$ & ${73.97} \pm 0.51$ \\
\noalign{\smallskip}
\hline
\noalign{\smallskip}
\end{tabular}
\end{center}
\end{table}
\setlength{\tabcolsep}{1.4pt}

\begin{figure}[!ht]
\centering
\includegraphics[width=\linewidth, trim = 0mm 125mm 75mm 1mm,clip]{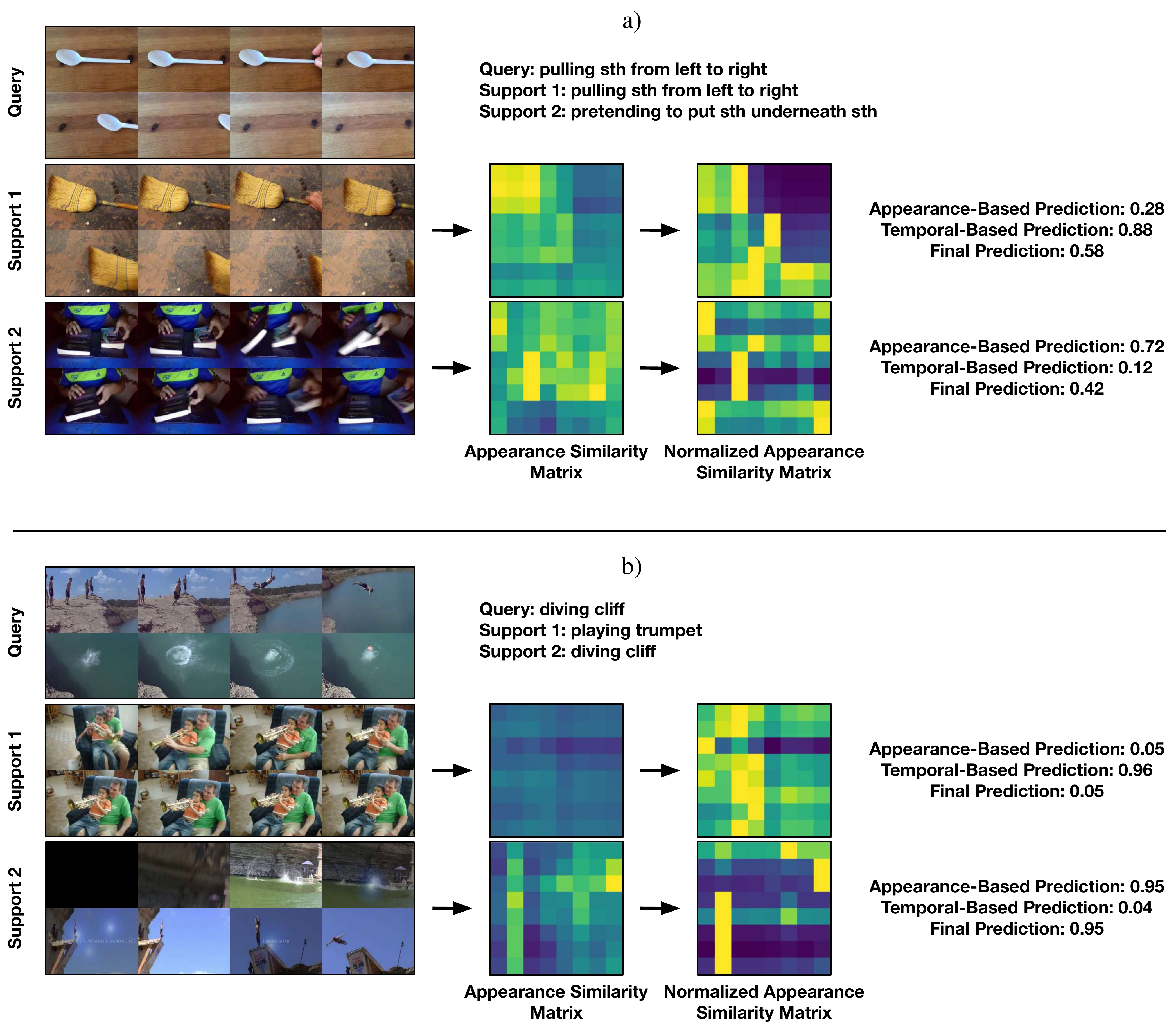}
\caption{\textbf{Qualitative Results of 2-Way 1-Shot Tasks on Something-Something V2 and Kinetics.} For each task, we present the appearance similarity matrix $\mathbf{D}$ between the query video and each support video in the second column. In the third column, we show the row-wise normalized version $\tilde{\mathbf{D}}$. Finally, we show the predictions of the two similarity scores and the final prediction.
Ground truth class labels are shown at the top.
a) Results from Something-Something V2. b) Results from Kinetics.}
\label{fig:example}
\end{figure}

The two table shows that for order-sensitive actions in Something-Something V2, balancing between the importance of appearance and temporal scores gives the best performance. In particular, for the inductive setting, our approach optimally achieves $48.55\%$ for 1-shot with $\beta=0.4$, whereas the best result for 5-shot is $66.91\%$ with $\beta=0.5$. Next, $\beta=0.3$ gives the best results for both 1-shot and 5-shot in the transductive setting, achieving $54.76\%$ and $75.47\%$ respectively.
Jointly considering appearance and temporal cues consistently improves the model performance as compared to the appearance-only version ($\beta = 0.0$). 
In contrast, temporal information does not show much benefits for an order-insensitive dataset like Kinetics. For small values of $\beta$, there are minor changes in the performance of our approach for both inductive and transductive settings on Kinetics. In the worst case, the temporal-only version ($\beta=1.0$) produces nearly random guesses of $21.09 \%$ in the 1-shot inductive setting.

We show some qualitative results in Fig.~\ref{fig:example}. We respectively perform inductive inferences on 2-way 1-shot tasks of Something-Something V2 and Kinetics datasets with $\beta=0.5$ and $\beta=0.0$ respectively.
On Something-Something V2 (Fig.~\ref{fig:example}(a)), we observe that appearance or temporal cues can misclassify query samples sometimes, but utilizing both appearance or temporal cues gives correct classifications. On the other hand, the actions on Kinetics ((Fig.~\ref{fig:example}(b)) are not order-sensitive, and hence the temporal similarity score is not meaningful, which agrees with the results in Tab.~\ref{table:abla:inductive}.

\setlength{\tabcolsep}{4pt}
\begin{table}
\begin{center}
\caption{Comparison to the state-of-the-art methods in the \textbf{inductive} setting on the Kinetics and Something-Something V2 datasets. Results of Meta-Baseline \cite{snell2017prototypical} and CMN \cite{zhu2018compound} are reported from \cite{zhu2021closer}. $\dag$ denotes results from our re-implementation.}
\label{table:inductive}
\begin{tabular}{c|c|c|c|c}
\hline\noalign{\smallskip}
 & \multicolumn{2}{c|}{\textbf{Kinetics}} & \multicolumn{2}{c}{\textbf{Something V2}} \\
\textbf{Method} & \textbf{1-shot} & \textbf{5-shot} & \textbf{1-shot} & \textbf{5-shot}\\
\noalign{\smallskip}
\hline
\noalign{\smallskip}
Meta-Baseline \cite{snell2017prototypical} & $ 64.03 \pm 0.41$ & $80.43  \pm 0.35$ & $37.31 \pm 0.41$ & $48.28  \pm 0.44$ \\
CMN \cite{zhu2018compound} & $65.90 \pm 0.42$ & $82.72 \pm 0.34$ & $40.62 \pm 0.42$ & $51.90 \pm 0.44$ \\
OTAM \cite{cao2020few} & ${73.00} \pm n/a$ & ${85.80} \pm n/a$ & $42.80 \pm n/a$ & $52.30 \pm n/a$ \\
Baseline Plus \cite{zhu2021closer}$^\dag$ & ${70.48} \pm 0.40$ & ${82.67} \pm 0.33$ & ${43.05} \pm 0.41$ & ${57.50} \pm 0.43$ \\
ITANet \cite{zhang2021learning} & ${73.60} \pm 0.20$ & ${84.30} \pm 0.30$ & $\bf{49.20} \pm 0.20$ & $\bf{62.30} \pm 0.30$ \\
\noalign{\smallskip}
\hline
\noalign{\smallskip}
\bf{Ours} & $\bf{74.26} \pm 0.38$ & $\bf{87.40} \pm 0.30$ & $43.82 \pm 0.42$ & $61.07 \pm 0.42$ \\
\noalign{\smallskip}
\hline
\end{tabular}
\end{center}
\end{table}
\setlength{\tabcolsep}{1.4pt}

\subsection{Comparison with Previous Methods}
\label{sec:comparison}
We compare our approach against previous methods~\cite{snell2017prototypical,zhu2018compound,ren2018meta,lichtenstein2020tafssl,cao2020few,zhang2021learning,zhu2021closer} in both inductive and transductive settings.
More specifically, in the inductive setting, we first consider Prototypical Network~\cite{snell2017prototypical} from few-shot image classification, which is re-implemented by~\cite{zhu2021closer}. In addition, CMN~\cite{zhu2018compound}, OTAM~\cite{cao2020few}, Baseline Plus~\cite{zhu2021closer}, and ITANet~\cite{zhang2021learning}, which are previous methods designed to tackle few-shot video classification, are also considered.
The results of Prototypical Network (namely, Meta-Baseline) and CMN are taken from~\cite{zhu2021closer}. We re-implement Baseline Plus~\cite{zhu2021closer}. 
The results of OTAM and ITANet are taken from the original papers. Competing methods in the transductive setting include clustering-based methods from few-shot image classification, namely Soft $K$-means \cite{ren2018meta}, Bayes $K$-means \cite{lichtenstein2020tafssl}, and Mean-shift \cite{lichtenstein2020tafssl}.

\noindent \textbf{Inductive.} We first consider the inductive setting. The results are presented in Tab.~\ref{table:inductive}. As can be seen from the table, our method achieves the best results on the Kinetics dataset. It outperforms all the competing methods by around $1\%$ for 1-shot setting and over $3\%$ for 5-shot setting, establishing the new state of the art. In addition, our method performs comparably with previous works on Something-Something V2, outperforming all the competing methods except for ITANet, which further adopts additional layers on top of the ResNet-50 for self-attention modules.

\noindent \textbf{Transductive.} Next, we consider the transductive setting (shown in Tab.~\ref{table:transductive}). 
As can be seen in Tab.~\ref{table:transductive}, our method outperforms all the competing methods by large margins, which are around $9\%$ for both 1-shot and 5-shot settings on Kinetics and $2\%$ and $5\%$ for 1-shot and 5-shot settings respectively on Something-Something V2.

\setlength{\tabcolsep}{4pt}
\begin{table}\label{tab:transductive}
\begin{center}
\caption{Comparison to the state-of-the-art methods in \textbf{transductive} setting on the Kinetics and Something-Something V2 datasets. Results of other methods are from our re-implementation on the trained feature extractor of \cite{zhu2021closer}.}
\label{table:transductive}
\resizebox{0.98\textwidth}{!}{
\begin{tabular}{c|c|c|c|c}
\hline\noalign{\smallskip}
 & \multicolumn{2}{c|}{\textbf{Kinetics}} & \multicolumn{2}{c}{\textbf{Something V2}} \\
\textbf{Method} & \textbf{1-shot} & \textbf{5-shot} & \textbf{1-shot} & \textbf{5-shot}\\
\noalign{\smallskip}
\hline
\noalign{\smallskip}
{Soft $K$-means}~\cite{ren2018meta} & $74.21 \pm 0.40$ & ${84.13} \pm 0.33$ & $46.46 \pm 0.46$ & $64.93 \pm 0.45$ \\
{Bayes $K$-means~\cite{lichtenstein2020tafssl}} & $70.66 \pm 0.40$ & $81.21 \pm 0.34$ & ${43.15} \pm 0.41$ & ${59.48} \pm 0.42$ \\
{Mean-shift~\cite{lichtenstein2020tafssl}} & ${70.52} \pm 0.40$ & ${82.31} \pm 0.34$ & ${43.15} \pm 0.41$ & ${60.03} \pm 0.43$ \\
\noalign{\smallskip}
\hline
\noalign{\smallskip}
\bf{Ours} & $\bf{83.08} \pm 0.44$ & $\bf{93.33} \pm 0.30$ & $\bf{48.67} \pm 0.54$ & $\bf{69.42} \pm 0.55$ \\
\noalign{\smallskip}
\hline
\noalign{\smallskip}
\end{tabular}}
\end{center}
\end{table}
\setlength{\tabcolsep}{1.4pt}

\section{Limitation Discussion}
We propose two similarity functions for aligning appearance and temporal cues of videos. While our results are promising in both inductive and transductive experiments, there remain some limitations. Firstly, our temporal order-preserving prior does not work for all datasets.
Utilizing a permutation-aware temporal prior~\cite{liu2021learning} would be an interesting next step. 
Secondly, we have not leveraged spatial information in our approach yet. Such spatial information could be important for scenarios like modeling left-right concepts. We leave this investigation as our future work. 

\section{Conclusion}
We propose, in this paper, a novel approach for few-shot video classification via appearance and temporal alignments. Specifically, our approach performs frame-level feature alignment to compute the appearance similarity score between the query and support videos, while utilizing temporal order-preserving priors to calculate the temporal similarity score between the videos. The proposed similarity scores are then used across different stages of our few-shot video classification framework, namely prototype-based training and testing, and inductive and transductive prototype enhancement. We show that our similarity scores are most effective on temporal order-sensitive datasets such as Something-Something V2, while our approach produces comparable or better results than previous few-shot video classification methods on both Kinetics and Something-Something V2 datasets. To the best of our knowledge, our work is the first to explore transductive few-shot video classification, which could facilitate more future works in this direction.

\appendix

\section{Supplementary Material}
In this supplementary material,  we first elaborate the difference between datasets that are sensitive or insensitive to action ordering in Sec.~\ref{A}. In Sec.~\ref{B}, we provide additional experiment results. We then evaluate the performance of our method on two action recognition benchmarks, namely UCF-101 and HMDB-51 in Sec.~\ref{C}.

\subsection{Order-Sensitive Datasets vs. Order-Insensitive Datasets}
\label{A}
\begin{figure}[!ht]
\centering
\includegraphics[width=\linewidth, trim = 0mm 2mm 30mm 1mm,clip]{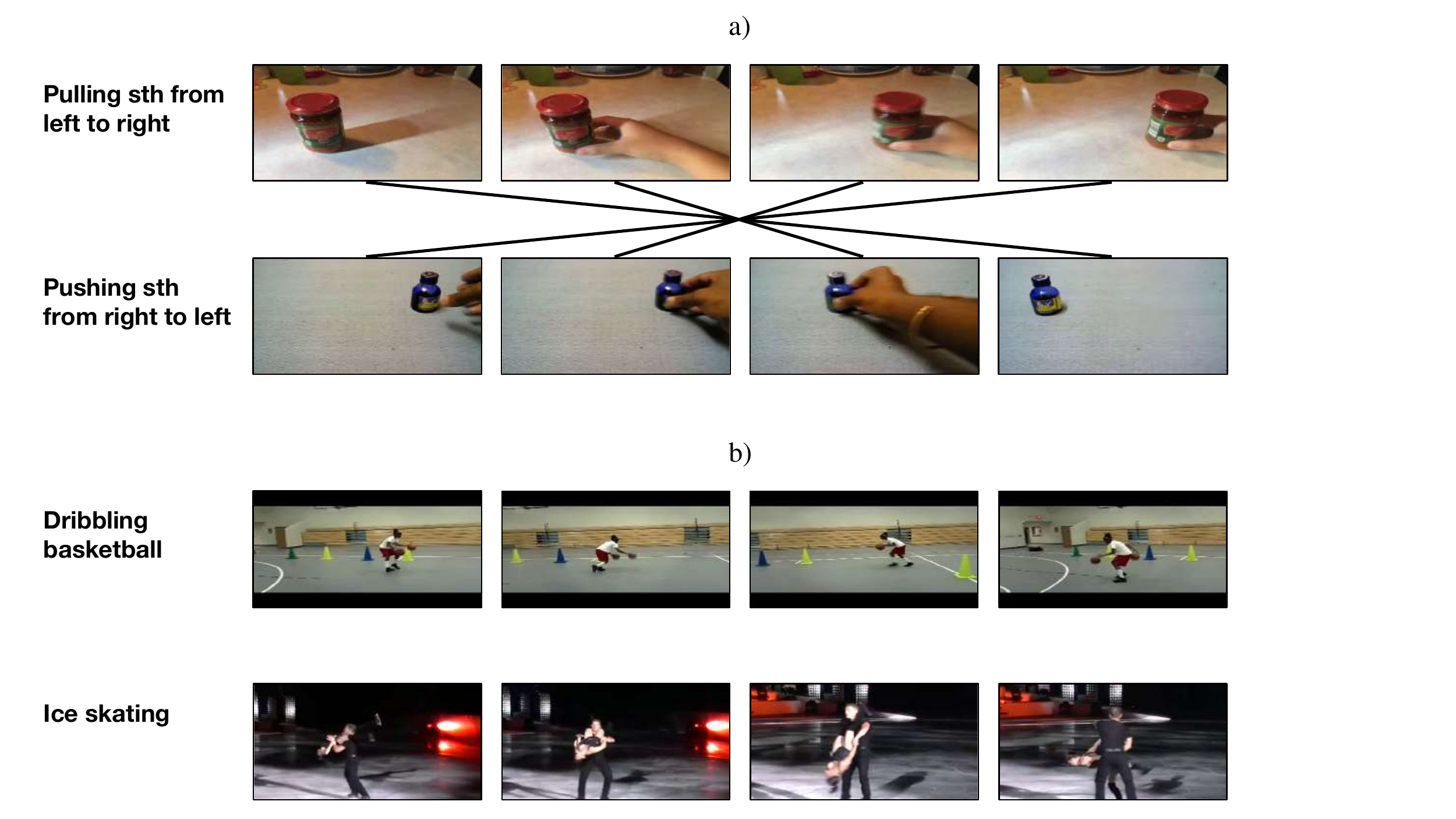}
\caption{\textbf{Action Ordering.} a) Example videos from Something-Something V2. The two videos can be aligned with each other in terms of appearance. To classify correctly, the model must consider the order of video frames. b) Example videos from Kinetics. Classes of Kinetics are mostly different in appearance and sometimes do not have a fixed frame order.}
\label{fig:sth_example}
\end{figure}

Here, we discuss the difference between datasets that are sensitive or insensitive to action ordering. For order-sensitive datasets, temporal cues (e.g., temporal order-preserving prior~\cite{su2017order,cao2020few,haresh2021learning,kumar2021unsupervised}) are essential in distinguishing between video categories. For example, in Something-Something V2, deciding whether a video belongs to "Pulling sth from left to right" or "Pushing sth from right to left" must consider the positional changes of an object presented in the video. Fig.~\ref{fig:sth_example} shows how videos from the two classes in Something-Something V2 can be aligned with each other. On the other hand, videos in order-insensitive datasets like Kinetics loosely rely on temporal cues. Their video classes can mostly be distinguished just with the appearance information.

\subsection{Additional Experiment Results}
\label{B}

\noindent \textbf{Run Times.} During training, we use the same ResNet-50 encoder with a linear classifier on top as the baseline~\cite{zhu2021closer}, yielding the same number of parameters. At inference stage, we discard the linear classifier and use the trained ResNet-50 to extract frame features. In our implementation, our method has the same number of epochs (25 epochs) and roughly the same training (12 hours) and inference time (0.2 and 0.5 secs for a 1-shot and 5-shot episode) as the baseline.

\begin{figure}[t]
	\centering
		\includegraphics[width=1.0\linewidth, trim = 20mm 175mm 0mm 51mm,clip]{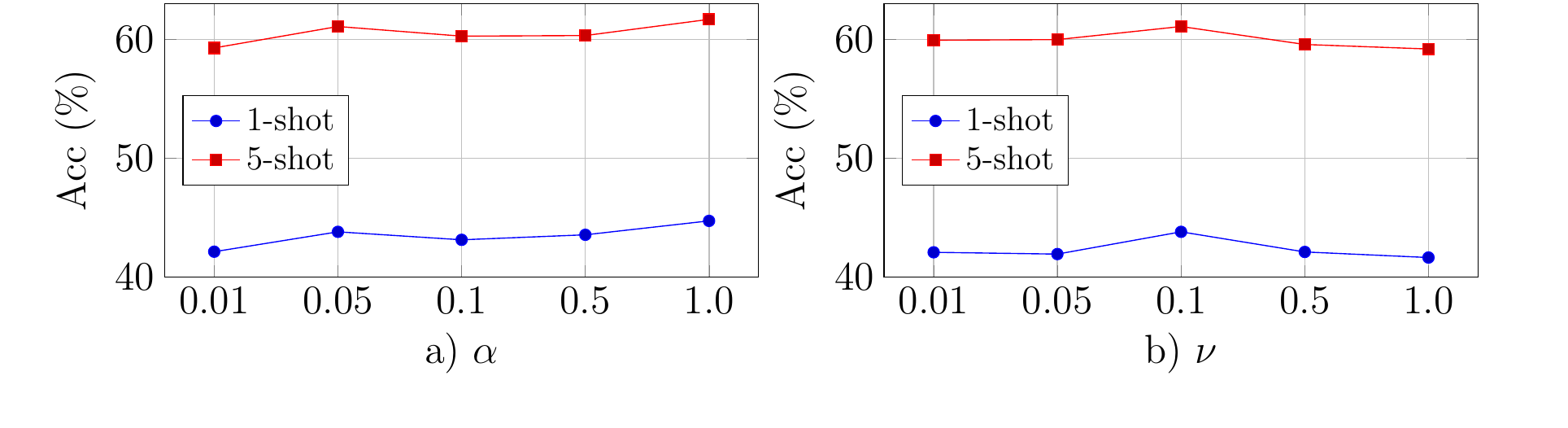}
	\caption{\textbf{Additional Ablation Results.} a) $\alpha$ and b) $\nu$.}
	\label{fig:rebuttal-ablation}
\end{figure}

\noindent \textbf{Additional Ablation Results.} 
We perform ablation study on the two hyperparameter $\alpha$ and $\nu$. We first fix $\nu = 0.1$ and vary the value of $\alpha$ in the range $[0.01, 1.0]$. Results are shown in Fig.~\ref{fig:rebuttal-ablation}(a). As we can observe, increasing the value of $\alpha$ improves the performance consistently.
We then fix $\alpha = 0.05$ and vary the value of $\nu$ in the range $[0.01, 1.0]$. Results are shown in Fig.~\ref{fig:rebuttal-ablation}(b).
$\nu = 0.1$ achieves the best results for both 1-shot and 5-shot settings.

\begin{figure}[h!]
\centering
\includegraphics[width=\linewidth, trim = 0mm 140mm 15mm 10mm,clip]{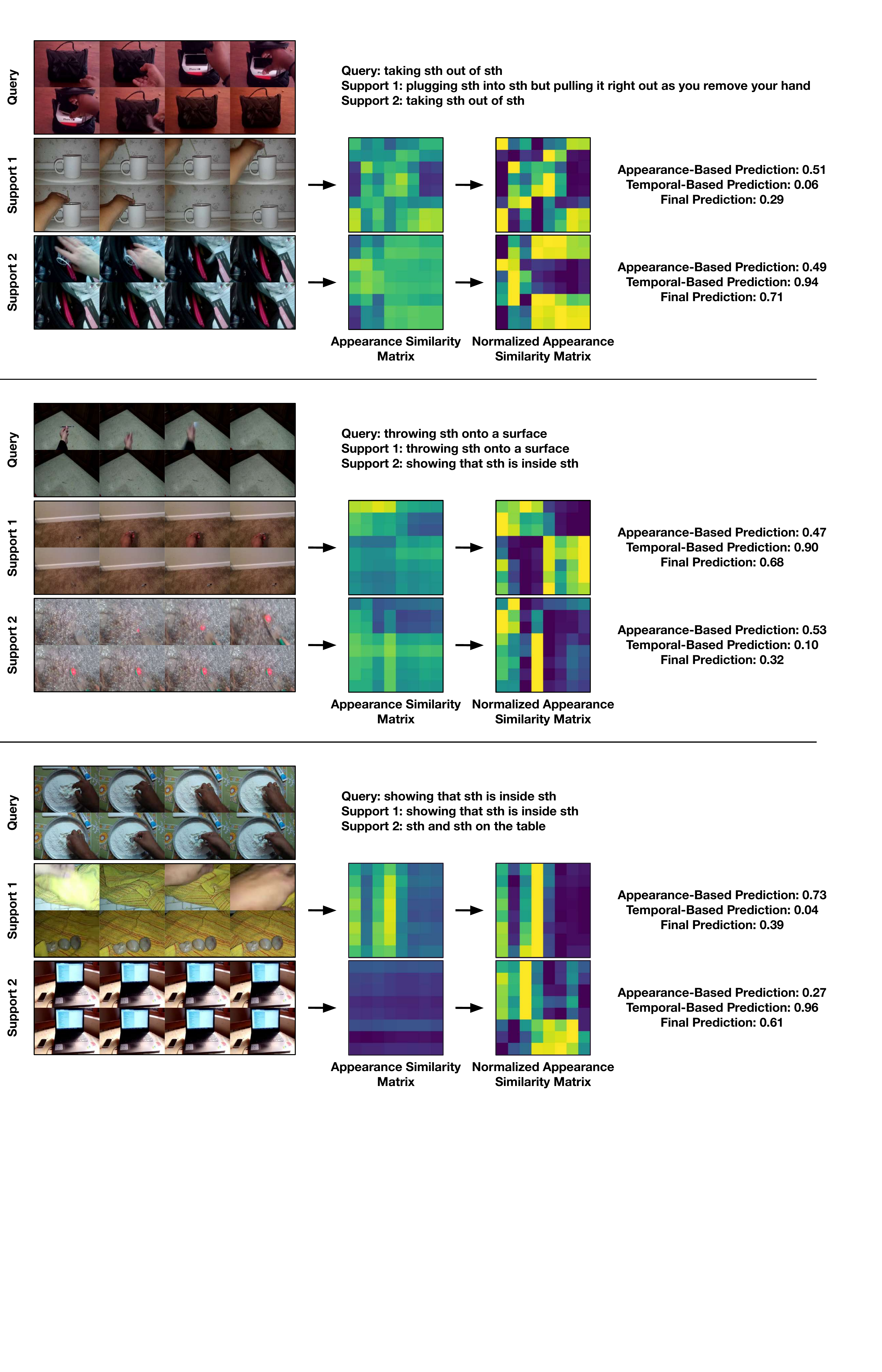}
\caption{\textbf{Qualitative Results of 2-Way 1-Shot Tasks on Something-Something V2.} For each task, we present the appearance similarity matrix $\mathbf{D}$ between the query video and each support video in the second column. In the third column, we show the row-wise normalized version $\tilde{\mathbf{D}}$. Finally, we show the predictions of the two similarity scores and the final prediction.
Ground truth class labels are shown at the top.}
\label{fig:supp-example}
\end{figure}

\noindent \textbf{Additional Qualitative Results.} 
We provide additional qualitative results of inductive inferences on 2-way 1-shot tasks of Something-Something V2 in Fig.~\ref{fig:supp-example}.

\subsection{Few-Shot Action Recognition Results on UCF-101 and HMDB-51}
\label{C}
So far we have focused on the problem of few-shot video classification. We now consider a related problem of few-shot action recognition~\cite{tan2019learning,kumar2019protogan,zhang2020few,cao2021few}. The main difference between video classification and action recognition is that, in video classification, videos/classes can describe general contents (e.g., ``sled dog racing'' in Kinetics), which are not limited to human actions as in action classification. In this section, we evaluate the performance of our method on two few-shot action recognition datasets, including UCF-101~\cite{soomro2012ucf101} and HMDB-51~\cite{kuehne2011hmdb}. UCF-101 is an action recognition dataset consisting of 13,320 YouTube videos of 101 action classes. 
For HMDB-51, there are 6,849 videos collected from different sources, i.e., movies, Prelinger Archive, YouTube, and Google videos. 
For both datasets, we follow the splits from \cite{zhang2020few}, which divide UCF-101 into 70/10/21 classes and HMDB-51 into 31/10/10 classes for training/validation/testing respectively.

\noindent \textbf{Implementation Details.} We apply the same preprocessing steps that we use for Kinetics and Something-Something V2 datasets to UCF-101 and HMDB-51 datasets. Specifically, a video is divided into $M=8$ segments and a frame is sampled randomly from each segment. We use the SGD optimizer with an initial learning rate of 0.0005 for both datasets. For UCF-101, the model is trained for 30 epochs, and we reduce the learning rate to $10^{-4}$, $10^{-5}$, and $10^{-6}$ at epochs 15, 20, and 25 respectively. For HMDB-51, the model is trained for 25 epochs, and the learning rate is reduced to $10^{-4}$, $10^{-5}$, and $10^{-6}$ at epochs 10, 15, and 20 respectively.

\subsection{The Relative Importance of Appearance and Temporal Cues on UCF-101 and HMDB-51}

We investigate the effectiveness of the hyperparameter $\beta$ (in Eq. 9 in the main paper) in the inductive and transductive settings on the UCF-101 and HMDB-51 datasets. Tabs.~\ref{table:sub:abla:inductive} and \ref{table:sub:abla:transductive} provide the mean accuracy with $95\%$ confidence interval on 10,000 episodes sampled from the validation set.

Tab.~\ref{table:sub:abla:inductive} shows that, in the inductive setting, appearance cues are more important than temporal cues for both datasets. In addition, utilizing both appearance and temporal cues ($\beta \in [0.2,0.4]$) yields minor improvements over using appearance cues only ($\beta = 0$). Similarly, for the transductive results in Tab.~\ref{table:sub:abla:transductive}, appearance cues play a more important role than temporal cues for both datasets. Moreover, leveraging both appearance and temporal cues ($\beta \in [0.2,0.4]$) leads to significant improvements on HMDB-51 but marginal performance gains on UCF-101, as compared to using appearance cues only ($\beta = 0$).


\setlength{\tabcolsep}{4pt}
\begin{table}[h]
\begin{center}
\caption{Ablation study on the relative importance of the appearance and temporal terms in computing the predictive distribution of the \textbf{inductive} inference on the UCF-101 and HMDB-51 datasets. Results are mean accuracy with $95\%$ confidence interval on 10,000 episodes sampled from validation set.}
\label{table:sub:abla:inductive}
\begin{tabular}{c|c|c|c|c}
\hline\noalign{\smallskip}
 & \multicolumn{2}{c|}{\textbf{UCF-101}} & \multicolumn{2}{c}{\textbf{HMDB-51}} \\
$\beta$  & \textbf{1-shot} & \textbf{5-shot} & \textbf{1-shot} & \textbf{5-shot}\\
\noalign{\smallskip}
\hline
\noalign{\smallskip}
$1.0$ & ${24.96} \pm 0.37$ & ${30.17} \pm 0.39$ & ${24.77} \pm 0.37$ & ${32.24} \pm 0.39$ \\
$0.8$ & ${35.32} \pm 0.41$ & ${80.24} \pm 0.37$ & ${30.13} \pm 0.39$ & ${53.14} \pm 0.43$ \\
$0.6$ & ${67.90} \pm 0.41$ & ${94.47} \pm 0.20$ & ${49.52} \pm 0.42$ & ${79.58} \pm 0.34$ \\
$0.4$ & ${84.55} \pm 0.31$ & $\bf{95.19} \pm 0.19$ & $\bf{65.78} \pm 0.38$ & $\bf{82.27} \pm 0.32$ \\
$0.2$ & $\bf{84.63} \pm 0.31$ & ${95.16} \pm 0.19$ & ${65.77} \pm 0.38$ & ${82.26} \pm 0.32$ \\
$0.0$ & ${84.60} \pm 0.31$ & ${95.15} \pm 0.19$ & ${65.72} \pm 0.38$ & ${82.21} \pm 0.32$ \\
\noalign{\smallskip}
\hline
\noalign{\smallskip}
\end{tabular}
\end{center}
\end{table}
\setlength{\tabcolsep}{1.4pt}

\setlength{\tabcolsep}{4pt}
\begin{table}[h]
\begin{center}
\caption{Ablation study on the relative importance of the appearance and temporal terms in computing the assignment function and the predictive distribution of the \textbf{transductive} inference on the UCF-101 and HMDB-51 datasets. Results are mean accuracy with $95\%$ confidence interval on 10,000 episodes sampled from validation set.}
\label{table:sub:abla:transductive}
\begin{tabular}{c|c|c|c|c}
\hline\noalign{\smallskip}
 & \multicolumn{2}{c|}{\textbf{UCF-101}} & \multicolumn{2}{c}{\textbf{HMDB-51}} \\
$\beta$  & \textbf{1-shot} & \textbf{5-shot} & \textbf{1-shot} & \textbf{5-shot}\\
\noalign{\smallskip}
\hline
\noalign{\smallskip}
$1.0$ & ${83.46} \pm 0.35$ & ${93.59} \pm 0.23$ & ${62.01} \pm 0.44$ & ${78.27} \pm 0.39$ \\
$0.8$ & ${88.95} \pm 0.32$ & ${97.34} \pm 0.17$ & ${70.26} \pm 0.46$ & ${87.70} \pm 0.35$ \\
$0.6$ & ${91.73} \pm 0.31$ & ${98.46} \pm 0.14$ & ${74.11} \pm 0.48$ & ${89.92} \pm 0.34$ \\
$0.4$ & ${92.72} \pm 0.30$ & ${98.74} \pm 0.13$ & ${75.08} \pm 0.49$ & $\bf{90.23} \pm 0.35$ \\
$0.2$ & $\bf{92.98} \pm 0.31$ & $\bf{98.75} \pm 0.13$ & $\bf{75.38} \pm 0.50$ & ${90.12} \pm 0.35$ \\
$0.0$ & ${92.91} \pm 0.31$ & ${98.68} \pm 0.14$ & ${75.02} \pm 0.49$ & ${89.39} \pm 0.36$ \\
\noalign{\smallskip}
\hline
\noalign{\smallskip}
\end{tabular}
\end{center}
\end{table}
\setlength{\tabcolsep}{1.4pt}

\subsection{Comparison with Previous Few-Shot Action Recognition Methods}

We compare our method with previous works on the two benchmarks UCF-101 and HMDB-51. 

\setlength{\tabcolsep}{4pt}
\begin{table}
\begin{center}
\caption{Comparison to the state-of-the-art methods in the \textbf{inductive} setting on the UCF-101 and HMDB-51 datasets. $\dag$ denotes results from our re-implementation.}
\label{table:sub:inductive}
\resizebox{0.98\textwidth}{!}{
\begin{tabular}{c|c|c|c|c|c|c}
\hline\noalign{\smallskip}
 & & \textbf{Pretrained} & \multicolumn{2}{c|}{\textbf{UCF-101}} & \multicolumn{2}{c}{\textbf{HMDB-51}} \\
\textbf{Method} & \textbf{Backbone} & \textbf{Dataset} & \textbf{1-shot} & \textbf{5-shot} & \textbf{1-shot} & \textbf{5-shot}\\
\noalign{\smallskip}
\hline
\noalign{\smallskip}
ProtoGAN~\cite{kumar2019protogan} & C3D~\cite{tran2015learning} & Sports1M~\cite{karpathy2014large} & ${61.70} \pm 1.60$ & ${79.70} \pm 0.80$ & $34.40 \pm 1.30$ & $50.90 \pm 0.60$\\
FAN~\cite{tan2019learning} & Densenet-121~\cite{huang2017densely} & - & ${71.80} \pm 0.10$ & ${86.50} \pm 0.20$ & $ 50.20 \pm 0.20$ & $67.60 \pm 0.10$\\
ARN~\cite{zhang2020few} & C3D~\cite{tran2015learning} & Sports1M~\cite{karpathy2014large} & ${62.10} \pm 1.00$ & ${ 84.80} \pm 0.80$ & $ 44.60 \pm 0.90$ & $59.10 \pm 0.80$\\
Baseline Plus~\cite{zhu2021closer}$^\dag$ & Resnet-50~\cite{he2016deep} & ImageNet~\cite{ILSVRC15} & ${81.06} \pm 0.33$ & ${92.85} \pm 0.22$ & ${57.56} \pm 0.42$ & ${72.70} \pm 0.37$ \\
ITA~\cite{cao2021few} & C3D~\cite{tran2015learning} & Kinetics-400~\cite{kay2017kinetics} & ${88.71} \pm 0.19$ & ${96.78} \pm 0.08$ & ${63.43} \pm 0.28$ & ${79.69} \pm 0.20$\\
CMOT~\cite{lu2021few} & C3D~\cite{tran2015learning} & Sports1M~\cite{karpathy2014large} & $\bf{90.40} \pm 0.40$ & $\bf{95.70} \pm 0.30$ & $\bf{66.90} \pm 0.50$ & $\bf{81.50} \pm 0.40$ \\
\noalign{\smallskip}
\hline
\noalign{\smallskip}
\bf{Ours} & Resnet-50~\cite{he2016deep} & ImageNet~\cite{ILSVRC15} & ${84.93} \pm 0.30$ & ${95.87} \pm 0.17$ & ${59.57} \pm 0.40$ & $76.85 \pm 0.36$\\
\noalign{\smallskip}
\hline
\end{tabular}
}
\end{center}
\end{table}
\setlength{\tabcolsep}{1.4pt}

\noindent \textbf{Inductive.} Inductive results are presented in Tab.~\ref{table:sub:inductive}. The competing methods include from few-shot action recognition approaches, i.e., FAN~\cite{tan2019learning}, ProtoGAN~\cite{kumar2019protogan}, ARN~\cite{zhang2020few}, and ITA~\cite{cao2021few}, as well as a recent few-shot video classification approach, namely Baseline Plus~\cite{zhu2021closer}. ARN, ProtoGAN, and ITA use pretrained C3D network as their backbone, while FAN use a pretrained Dense-121 backbone network. Their results are taken from the original papers. The results of Baseline Plus are from our re-implementation. The datasets used for backbone network pretraining are presented in Tab.~\ref{table:sub:inductive}, except for FAN which does not mention that detail in their paper. Tab.~\ref{table:sub:inductive} shows that CMOT performs the best, followed by ITA, which achieves the second best performance across all settings on both datasets. However, we note that CMOT and ITA use a C3D backbone, which is capable of extracting spatiotemporal information, and pretrained on a video dataset (i.e., Sports1M, Kinetics-400). In contrast, we use a 2D-based backbone (i.e., ResNet-50), and pretrain it on ImageNet dataset.

\setlength{\tabcolsep}{4pt}
\begin{table}
\begin{center}
\caption{Comparison to the state-of-the-art methods in \textbf{transductive} setting on the UCF-101 and HMDB-51 datasets. Results of other methods are from our re-implementation on the trained feature extractor of \cite{zhu2021closer}.}
\label{table:sub:transductive}
\begin{tabular}{c|c|c|c|c}
\hline\noalign{\smallskip}
 & \multicolumn{2}{c|}{\textbf{UCF-101}} & \multicolumn{2}{c}{\textbf{HMDB-51}} \\
\textbf{Method} & \textbf{1-shot} & \textbf{5-shot} & \textbf{1-shot} & \textbf{5-shot}\\
\noalign{\smallskip}
\hline
\noalign{\smallskip}
{Soft $K$-means}~\cite{ren2018meta} & ${90.26} \pm 0.34$ & ${97.67} \pm 0.17$ & ${64.44} \pm 0.52$ & ${80.29} \pm 0.45$ \\
{Bayes $K$-means~\cite{lichtenstein2020tafssl}} & ${81.26} \pm 0.33$ & ${93.13} \pm 0.22$ & ${57.85} \pm 0.42$ & ${73.44} \pm 0.37$ \\
{Mean-shift~\cite{lichtenstein2020tafssl}} & ${81.25} \pm 0.33$ & ${89.35} \pm 0.26$ & ${57.82} \pm 0.42$ & ${67.96} \pm 0.43$ \\
\noalign{\smallskip}
\hline
\noalign{\smallskip}
\bf{Ours} & $\bf{94.18} \pm 0.28$ & $\bf{99.06} \pm 0.11$ & $\bf{68.07} \pm 0.52$ & $\bf{85.01} \pm 0.42$\\
\noalign{\smallskip}
\hline
\noalign{\smallskip}
\end{tabular}
\end{center}
\end{table}
\setlength{\tabcolsep}{1.4pt}

\noindent \textbf{Transductive.} 
Next, we consider the transductive setting. We re-implement three transductive techniques from few-shot image classification, i.e., Soft $K$-means~\cite{ren2018meta}, Bayes $K$-means~\cite{lichtenstein2020tafssl}, Mean-shift~\cite{lichtenstein2020tafssl} as our competing methods. Results are shown in Tab.~\ref{table:sub:transductive}. To our best knowledge, we are the first to consider transductive inference in few-shot action recognition. As it is evident from the results, our method significantly outperforms the competing methods by large margins, i.e., $2-4\%$ on UCF-101 and $4-5\%$ on HMDB-51.

%
%
\bibliographystyle{splncs04}
\bibliography{egbib}

\end{document}